# Title: Counterfactual rewards promote collective transport using individually controlled swarm microrobots

**Authors:**

Veit-Lorenz Heuthe[1,2], Emanuele Panizon[3,4], Hongri Gu[1*], Clemens Bechinger[1,2*]

**Affiliations:**

[1] University of Konstanz, Department of Physics, Universtaetsstrasse 10, Konstanz, 78464, Germany

[2] Centre for the Advanced Study of Collective Behaviour, Universtaetsstrasse 10, Konstanz, 78464, Germany

[3] The Abdus Salam International Centre for Theoretical Physics (ICTP), Strada Costiera 11 Trieste, 34151, Italy

[4] Area Science Park, Data Engineering Laboratory, Località Padriciano 99, Trieste, 34149, Italy

*corresponding authors. email: hongri.gu@uni-konstanz.de (H.G.); clemens.bechinger@uni-konstanz.de (C.B.)

**Abstract:** Swarm robots offer fascinating opportunities to perform complex tasks beyond the capabilities of individual machines. Just as a swarm of ants collectively moves large objects, similar functions can emerge within a group of robots through individual strategies on the basis of local sensing. However, realizing collective functions with individually controlled microrobots is particularly challenging because of their micrometer size, large number of degrees of freedom, strong thermal noise relative to the propulsion speed and complex physical coupling between neighboring microrobots. Here, we implement Multi-Agent Reinforcement Learning (MARL) to generate a control strategy for up to 200 microrobots whose motions are individually controlled by laser spots. During the learning process, we employ so-called counterfactual rewards that automatically assign credit to the individual microrobots, which allows fast and unbiased training. With the help of this efficient reward scheme, swarm microrobots learn to collectively transport a large cargo object to an arbitrary position and orientation, similar to ant swarms. We show that this flexible and versatile swarm robotic system is robust to variations in group size, the presence of malfunctioning units, and environmental noise. Additionally, we let the robot swarm manipulate multiple objects simultaneously in a demonstration experiment, highlighting the benefits of distributed control and independent microrobot motion. Control strategies such as ours can potentially enable complex and automated assembly of mobile micromachines, programmable drug delivery capsules, and other advanced lab-on-a-chip applications.

**One-Sentence Summary:** Multi-agent RL with counterfactual rewards controls swarm microrobots to rotate, transport, and control objects in parallel.

# Main Text:

## Introduction

Many animals organize into swarms and develop cooperative strategies to solve problems and gain advantages beyond their individual capabilities. Examples of such collective achievements include the coordination of hunting strategies in pods of dolphins (*1*), the construction of nests by termites (*2*), and the collective transport of heavy objects by ants or social spiders (*3, 4*). Researchers in robotics are taking inspiration from these natural systems and are designing swarms of relatively simple robots that can perform complex tasks by working together (*5, 6*). At the group level, swarm robotic systems naturally possess several strengths, such as flexibility, scalability, and robustness. These advantages are a result of the distributed control and simplicity of the robots in a robot swarm. Without changing the hardware, the same robotic swarm can be used for different tasks by reprogramming the reactions of each robot to its neighbors and the detected environment. Similarly, the system size can be altered by adding more copies of the same unspecific unit to it. Swarm robotic systems tend to be robust since there is no central unit whose failure is critical and malfunctioning units can be swiftly replaced by neighbors. To date, many types of swarm robotic systems exist in which individual robots work collectively, opening the door to many exciting future applications in the air, on land, and at sea (*7-9*). Examples include miniaturized quadcopters that can avoid collisions and explore an unknown crowded environment (*10*), motorized cubes that can be reconfigured into programmable shapes (*11*), termite-like wheeled robots that can carry bricks and build castles (*12*), robotic fish swarms that exhibit complex, decentralized 3D collective behaviors through light communication (*13*) and swarms of buoys that can monitor a prescribed area (*14*). Within these swarm systems, each robot has a programmable controller that can make independent decisions on the basis of locally acquired information, enabling complex collective functions comparable to those we see in nature.

However, scaling these independently controlled swarm robots down to the micrometer range presents substantial challenges. It is often prohibitively challenging to integrate sensors, microcontrollers, microactuators, and batteries at such a small scale (*15, 16*). Instead, researchers are exploring the control of microrobotic swarms by means of a global field (e.g., electric, magnetic, or acoustic) for the entire workspace. Many impressive collective behaviors have been achieved by varying only a few system parameters. For example, the collective formation of swarms, chains, clusters, and ripples of charged or magnetic colloids is possible by tuning the electric or magnetic field signal (*17-19*). Using a dynamically alternating magnetic field, millions of magnetic nanoparticles can change the shape of a densely packed active swarm and navigate in a maze (*20*). Other works control bacteria to employ them as microrobot swarms (*21*). In these systems, the dominant local interactions (e.g., magnetic dipole–dipole interactions) are responsible for collective behaviors. Compared with swarms of macroscopic robots, where interactions can be individually controlled and programmed with almost arbitrary spatiotemporal rules, the local interactions between microrobots controlled by a global field are limited to simple physical interactions, resulting in a limited number of collective motion patterns. Control fields that extend over large distances also do not easily allow different tasks to be executed simultaneously and independently of each other. This suggests that the true potential of functional and intelligent behaviors in microrobotic swarms remains largely unexplored.

To independently control individual microrobots in a swarm, drive signals must be pinpointed locally with very high precision. Owing to their high spatial resolution and accuracy, lasers provide a powerful tool for selectively controlling individual microrobots. By programming structured light patterns, various types of microrobots, including mass-produced electronically integrated microbots (*22*), genetically modified C. Elegans (*23*), eight-legged microcrawlers (*24*), shape-memory alloy microrobots (*25*), ThermalBots at an air–water interface (*26*), and carbon-capped Janus microswimmers (*27*), can all be controlled by lasers to move independently. These kinds of microagents could be used to perform on-chip manipulation and assembly tasks, where the local actuation would allow for executing different tasks on a small space in parallel. Despite this powerful driving mechanism, many fundamental challenges remain. The first challenge is the presence of thermal noise, which has a large influence on micron-sized robots. When the Brownian motion is on the same order of magnitude as the locomotion speed, it is very difficult to precisely control the microrobots to follow a predetermined trajectory. This means that control algorithms must address uncertainties so that the collective functions are robust against these local disturbances. The second challenge is the strong and unpredictable physical coupling between multiple microrobots (*28*). For example, when a robot is hit by a focused laser beam, a large amount of energy is pumped into the local environment, changing the temperature, chemical concentration, and other physical properties associated with the propulsion mechanism. This drastic change in the microenvironment alters the interactions between microrobots in close contact and may disturb their motion. Simplified models usually fail to accurately capture such complex effects. Although direct contact between microrobots is often intentionally avoided because of such effects, many applications, such as cargo transport and collective microassembly, require microrobots to directly push against each other or other objects (*29, 30*).

Multi-Agent Reinforcement Learning (MARL) is an emerging field of artificial intelligence that offers a promising solution to complex control problems for swarm microrobots (*31*). In MARL, the control strategy (called policy) for the individual robots (called agents) is automatically optimized for a specific task without the need for an accurate model (*32*). This is achieved during a training procedure, where the agents actively explore their environment through observed states and subsequent actions. After each action, the performance of each agent with respect to the task is evaluated by a quantity called the reward (Fig. 1). An optimization algorithm is used to update the policy to favor actions associated with higher rewards in the past and to avoid actions that yielded lower rewards. In most cases, updating the policy involves altering the parameters of a neural network. Using MARL, the challenge of designing a complicated control algorithm for swarm robots is reduced to finding an appropriate reward function that quantifies the performance of the robots in their task. All system-specific details are then automatically handled by the learning algorithm, and there is no need for the controller to consider task-specific environmental conditions. However, writing an effective reward function for the individual agent in a multi-agent system is often challenging. Simply giving an identical team reward to all agents can lead to the 'lazy agent problem', where some agents do not contribute much but are still rewarded for others' work, undermining collective performance (*33*). Additionally, the variance in the reward signal of each robot is usually very high when team rewards are used, resulting in long training times and poor performance. For these reasons, it is important to formulate rewards that consider the contribution of each individual agent to group performance. However, designing individual reward functions for a specific task manually requires a very good understanding of the relationship between individual behavior and collective performance in this exact task. For this type of manual reward scheme, it is necessary to redesign the reward function for each task, which reduces the flexibility of the swarm robotic system. The method of counterfactual rewards provides a simple and convenient way to assign rewards to individual agents by comparing the performance in each time step in a real experiment to that in a hypothetical scenario. In these virtual copies of the system, one agent is removed, and the collective performance of the rest of the swarm is subsequently extrapolated for a short time step. By comparing the hypothetical result where one agent is removed with the actual experiment, the contribution of this specific agent can be determined and assigned as the individual reward for MARL. One disadvantage of counterfactual rewards is that simulating hypothetical scenarios requires a model of the system. Since the extrapolation step is rather small, however, the model does not need to be very accurate. Counterfactual rewards, as a method for generating individual rewards, as well as other similar credit assignment methods, are widely acknowledged in MARL but have not yet gained popularity in the swarm robotics community.

In this study, we developed a swarm microrobotic system utilizing up to two hundred individually controlled microrobots. Each microrobot is designed as a Janus microparticle (a 6 µm silica sphere with a carbon cap on one half) and is propelled by a laser spot. In our system, each robot has a discrete set of motion options (forward movement, left turns, right turns, and remaining stationary), as illustrated in Fig. 1C. The propulsion speed typically averages one tenth of the robot diameter per second, which is comparable to Brownian motion. We employ a MARL algorithm to train a control policy of the individual robots for the collaborative manipulation and transportation of a large rod-shaped cargo. As a policy, we use an artificial neural network (see the ANN section in the Supplementary Materials for architecture details). We introduce counterfactual rewards to assign credits to individual microrobots on the basis of their contribution to the collective performance, effectively mitigating the "lazy agent problem" and the high variance in rewards connected to team rewards. Our results demonstrate that MARL, in conjunction with the counterfactual reward approach, can acquire effective strategies for swarm control at the microscale using very restricted information obtained by each robot locally. This control strategy effectively overcomes the inherent thermal noise and intricate interactions between agents and cargo, including direct collisions and complex surface interactions. Our findings provide concrete evidence for the efficacy of counterfactual reward-enabled MARL in individually controlled swarm microrobots, extending the realm of swarm intelligence and collective functions to the micrometer scale. Movie 1 shows a summary of this study.

# Results

**Experimental realization**

In this study, we realized a microrobotic swarm system composed of laser-activated microswimmers (microrobots) that move in a two-dimensional plane on the bottom of a glass cell. The microswimmers were made from silica microspheres that were 6 µm in diameter and were coated with a light-absorbing 80 nm carbon cap on one hemisphere (Figure S1 shows an image of such a robot). When such particles are immersed in a binary fluid mixture (e.g., water and lutidine) close to its critical point, illumination with a focused laser beam leads to selective heating of the carbon cap and thus to local demixing of the fluid. The asymmetric demixing of the fluid around a particle creates a force on its surface that makes it move (*34*). Depending on the intensity and position of the laser focus relative to the carbon cap, the robot can be steered to move forward and even turn. Using a feedback loop and live tracking together with an acousto-optical deflector, we scanned a laser over several hundred particles with a high frequency and thus controlled the motion of each particle simultaneously (*34, 35*). We scanned at 100 kHz, which is much faster than the relaxation time of the propulsion mechanism and can access an area of 600 µm by 700 µm (for more details, see Materials and Methods and Fig. S2) (*36*). The average laser power applied to one microrobot in our experiment ranged from 0.3 µW to 2.7 µW. We let these microrobots interact with a rigid 6 x 6 x 100 µm ellipsoidal rod particle, which was fabricated via two-photon polymerization 3D printing (*37*).

To enable the swarm to perform a specific task, e.g., collectively moving an object, each microrobot must constantly adjust its motion according to the configuration of its surroundings. We achieved this by treating each microrobot as a MARL agent that can sense its environment and then decide about its next action. As actions, we allowed the robots to move forward, turn left, turn right or remain stationary. The decision for each microrobot was computed via a neural network fed with the partial information this robot has access to: In our experiments, each agent could measure the presence of other microrobots only by detecting the crowdedness in five equal detection cones with an angle of 36° to its front (yielding a 180° vision field; see Fig. 2A). For each cone, the robot received a state input given by the number of robots in this cone inversely weighted with the distance to these robots (see State input definition section in Materials and Methods). The rods were represented as a linear chain of circular particles in a second set of state inputs (Fig. 2B). In this way, each microrobot received a total of ten state inputs, five for other robots and five for the rod, as a representation of the current state of the system. Since we did not include any communication between the microrobots, the available information is very limited, and no agent had global information about the system. Because of this, our approach can be immediately applied to other systems in which no global information is available, even though our experiments were operated by a central computer. For many envisioned applications (e.g., parallel on-chip testing and manipulation, search and rescue scenarios in hazardous), centralized control with global information cannot be implemented, but robotic swarms are still required to coordinate with each other, and we believe that MARL holds great promise in solving these challenges.

The experimental state inputs were directly fed into the policy in the form of a neural network from which each microrobot received its next action. Using a MARL framework, we optimized this policy toward different tasks, which are explained in detail below (see Supplementary Materials for a detailed description of our learning algorithm). This optimization relies on a reward function that quantifies the performance of each microrobot. In contrast to simple tasks, which do not require cooperative behavior, the individual assignment of rewards in the context of a collective task is not straightforward: since the group's performance is defined only on a global scale, it is not immediately obvious how to acknowledge the contribution of a single agent to the overall task. A typical approach is to simply give every agent the same reward on the basis of the performance of the whole group, the so-called team reward. This, however, can lead to situations in which only some agents actively pursue the task and others are "lazy" but still receive rewards for the performance of the active agents (*38*). Additionally, in a MARL task such as ours, where the agents have only partial information about the system, an agent might be unable to differentiate between situations where it is rewarded owing to its own actions and those where it is rewarded because of the other agents' actions, which are outside its perception region. This makes it difficult for the agent to learn how to contribute to the performance of the group (*33*). To circumvent this general problem of credit assignment, we used counterfactual rewards (CRs) that credit each agent for its own contribution to the collective performance (*39*). In brief, our implementation of CR compares the group performance with and without a specific agent and estimates the agent's contributions from the corresponding performance difference. For the implementation of CR in our robotic swarm, we compared the experimentally measured performance with the simulated performance under the exclusion of one microrobot. Since the performance difference was only calculated over the time interval of a single action, a simple extrapolation of the robot's trajectories using an overdamped Langevin simulation provides a good estimate of the CR (see Supplementary Materials). For these re-simulation steps, we did not add any noise to the virtual environment because if both the experiment and the virtual environment are subject to noise, the difference in the outcome between the two is greater than when the noisy results of the experiment are compared to the noiseless result of the re-simulations. This individual reward scheme for microrobots within a swarm can be applied to a large range of collective tasks by simply defining different global measures of performance. We note that, like in any other reward scheme in RL, CRs are required only during training of the agents until the policy has been optimized but not for deploying the trained system. The Supplementary Materials contain a detailed comparison of CRs with team rewards and one realization of manually designed individual rewards (section Reward comparison). In short, we find that models trained with CRs do indeed exceed team rewards in terms of training speed and final performance. The models trained with manually designed rewards suffer from a biased strategy that leads to lower performance as well.

**Microrobot motion**

As outlined above, the robots could choose to move forward, rotate right, rotate left or stay stationary. The selected action was then applied to the microrobot via the laser spot pattern for the next 10 seconds. During this time, the laser spot positions were updated relative to the tracked microrobot position with 3 fps. Fig. 3A-E shows the resulting robot motion for the four different actions. For the "stationary" actions, a weak laser spot was applied to the back side of the carbon cap. This approach has been proven to reduce the orientational diffusion of the robots. During forward actions, a laser beam was also positioned at the back of the carbon cap of the microrobot but with higher intensity. This makes the robot propel in the direction of its orientation. When the robot has traveled 6 μm (one diameter), the "stationary" action is applied for the remaining 10 seconds of the current action. Continuously choosing the forward action allows the microrobot to move on a more or less straight trajectory. Brownian motion, however, leads to diffusion of the robot's orientation and introduces random curvature (see Supplementary Movie S1). In extreme cases, this noise can even turn the orientation of a robot by 180°, since we only measure the orientation in the measurement plane, but the robots can still rotate out of plane. When a robot chose a turning action, two laser spots with different intensities were applied to either side of the carbon cap. As a simplified explanation, more intense spots lead to stronger propulsion on one side, allowing the robot to swim on a curve. During rotation actions, the microrobots can rotate by 36°. Like in forward actions, the action was switched to "stationary" when the microrobot had reached the prescribed orientation. While rotating, the robots do not exactly retain their position but are displaced by the radius of a microrobot (3 μm) during one action. The direction of this translation is highly variable but makes the robot move forward on average, resulting in a curved trajectory as the combination of rotation and translation (as depicted in Fig. 3 and in Supplementary Movie S1). Even though action selection occurs in discrete timesteps, a microrobot can move on very complex trajectories because the distance traveled during each action is small.

Both the orientations and positions of the microrobots are strongly influenced by thermal noise. The diffusion coefficient of the microrobots is 0.03 µm/s, which translates to motion of approximately one tenth of the robot diameter during one action. In addition to Brownian noise, the interactions between the robots and the rod introduce even more noise to the robots' trajectories. As depicted in Fig. 3G, the swimmers often stick together when they collide until their propulsion allows them to move apart again. We observed similar interactions between the robots and the rods. In both cases, the maneuverability of the robots in contact is highly restricted (see Fig. S3 and Supplementary Movie S2). This manifested in broader distributions of displacement and rotation for all four actions. The sticky interaction is most likely caused by the demixing bubbles that propel the robots: when the bubbles of different robots overlap or touch the rod, phoretic interactions result in an attraction between the robots or a robot and the rod. Owing to the high complexity of the underlying mechanism, the exact effect of collisions on the motion of microrobots is very difficult to predict (*40, 41*). Since the robots must push against the rod to move it, these imperfections are not occasional occurrences but rather an important part of the problem the robot swarm faces. The MARL algorithm therefore needs to be able to address these uncertainties and enable the robots to develop behaviors that are robust enough to still carry out collective functions.

**Rotation of an ellipsoid rod**

The first task we considered was the rotation of a rod (6 × 6 × 100 µm ellipsoid). Owing to the large size and high fluidic drag of the rod, a single microrobot cannot effectively move it. Instead, cooperation and collective motion of the microrobotic swarm are required to push the rod. As a measure of swarm performance, we considered the absolute value of the angular velocity $\omega$ of the rod, and individual rewards for each microrobot were allocated via CRs. Training was performed in an episodic fashion with swarms of 30-35 robots, which were pushed onto a square grid around the rod at the beginning of each episode (see Fig. S4 and Supplementary Movie S3).

We found that during the first episodes, the microrobots move around randomly. This is expected since the neural network was initiated with random weights and thus produces random decisions before training. However, some robots collided with the rod and rotated it by chance, for which they were rewarded. After approximately ten episodes, we observed that the microrobots had learned that it was good to interact with the rod and try to navigate toward it. This is reflected in the reduced average distance between the robots and the rod (Fig. 4C). However, the angular velocity of the rod remained close to zero until the 20th episode since the robots still encounter the rod in an uncoordinated fashion. Fig. 4E shows the distribution of the robots along the rod length, which is still symmetric during that time. From the 20th episode on, we found an increase in the angular velocity of the rod. The robots achieved this by starting to systematically push against the ends in the clockwise direction, in contrast to their previous uncoordinated encounters. We evaluated the geometric torque

$$T_{geom} = \left| \sum_{i}^{N} \boldsymbol{r}_i \times \boldsymbol{u}_i \right| \quad (1)$$

(with the robot position with respect to the rod $\boldsymbol{r}_i$ and the robot orientation $\boldsymbol{u}_i$) over time, which reflects the onset of coordinated efforts of the microrobots (Fig. 4D). From episodes 20 to 40, the angular velocity of the rod rapidly increased and eventually saturated, whereas the average distance between the robots and the rod continued to decrease, and the geometric torque continued to increase (see Fig. 4D). The mean reward of the group (Fig. S5) showed the same evolution as the angular velocity. We conclude that counterfactual rewards provide an accurate measure of the individual performance of the microrobots, despite the simplification of the system in the numerical representation. Fig. S6 shows the learning curves for more experimental training runs. Supplementary Movie S3 shows experimental videos during different stages of training. Since we rewarded the microrobots on the basis of the absolute value of the angular velocity of the rod, the swarm adapts one sense of rotation at random, and we find that different trainings result in different senses of rotation. A trained model does, however, produce the same direction of rotation in every episode.

The first task of rotating the rod demonstrates that our system has several qualities that are necessary for a microscopic robot swarm. We achieved this result in end-to-end training, where the robots had to explore the real, physical environment to learn how to gain rewards and solve the task. The partial information and limited action space available to each robot were sufficient for them to navigate their surroundings and to accumulate at the rod. There, the microrobots managed to coordinate their actions to collectively push against the ends of the rod and rotate it. Despite the high level of noise in the individual robots' trajectories and the uncertainties introduced by collisions between the microrobots and the rod, all of these steps are possible, demonstrating the robustness of the MARL algorithm.

Notably, the counterfactual reward scheme demonstrated here does not require any a priori assumption regarding the specific microrobot strategy for rod rotation. In simulations, we compared the CRs with one instance of a manually designed reward function on the basis of the torque each microrobot exerts on the rod (see Reward comparison section in the Supplementary Materials). Training with torque-based rewards led to impaired final performance and biased strategies. Apart from performance deficits, manually designed reward functions rely on a priori assumptions about the strategy for solving a specific task, which are not obvious for most tasks. Using CRs, we were able to simply exchange the measure of global performance in the CR computation and thus train the microrobots to transport the rod to a defined target position, as discussed in the next section.

**Targeted transport**

As a second example of a collective task, we considered the transport of the rod (*42*) to a predefined arbitrary position and orientation (target area). For this task, we added another state input for each detection cone, which enables the robots to perceive the target area (similar to the rod; see State input definition section in the Materials and Methods for details). In addition, we expanded the total detection field to 360° by adding five more cones pointing to the back of each robot. This allowed the microrobots to simultaneously detect the rod and the target area independent of the robot orientation. To quantify the group performance, we divided the rod and the target into 60 identical segments each. Since we wanted to reward the robots for reducing the rod–target distance, we used the negative change in the mean pairwise distance between those segments as a measure of performance (see Fig. 5B and Reward definition section in Materials and Methods). Individual rewards were again obtained by assigning contributions to single robots via CRs. When the robots succeeded in moving the rod into the target site within one episode, an additional reward was assigned to all the robots to enhance the convergence of the optimization. Each episode started with a random orientation $\alpha$ of the rod and a center-to-center distance to the target of $D$ = 50 µm, which was placed in a random direction. Owing to the random placement of the target, the initial configuration was different for each episode. The targeted transport task is more complex than the rotation task since, first, all three modes of moving the rod (rotation and translation in both directions) are involved, and second, the exact rod motion necessary to fulfill the task varies from episode to episode owing to random initializations. This increase in complexity led to longer training times, which were unfeasible for end-to-end training. As a consequence, we trained the robot swarm in the virtual environment and subsequently applied the optimized policy to the experimental system. Although not optimal due to the simplification of the training environment, this zero-shot approach led the microrobots to push the rod to the target area with a success rate of more than 90 % within episodes of fewer than 3000 actions (as evaluated after training, see Supplementary Movie S4). One major difference in the behavior of the microrobots compared with the rotation task was the lower number of robots that are simultaneously engaging with the rod (see Supplementary Movie S5). We attribute this difference to the noisy and chaotic behavior of the robots in the experiment (see Fig. 3 and Fig. S3), which is not fully captured in the simulation in which the robots were trained. We attempted to increase the performance of the microrobot swarm and the engagement of the microrobots by continuing training in the experiment. However, additional training did not increase the performance (see Fig. S10). The performance of the microrobot swarm could be improved by implementing a more accurate model of individual robot behavior in the virtual environment. Such detailed models are, however, usually not available for microrobotic systems or other robot swarms in a priori unknown environments. The success of the zero-shot transfer of trained models from simulation to experiment shows that the MARL framework with counterfactual rewards even yields functioning policies in this typical case of a complex environment for which no detailed model exists.

We can identify three distinct sub-strategies that the microrobot swarm has adopted after training to solve the task. Each sub-strategy moved the rod in one of its degrees of freedom, namely, rotation, longitudinal transport and translational transport (see Fig. 5C). To rotate the rod around its center, the robots developed the same method as in the rotation task and pushed against its ends from opposite sides. Transporting the rod transversally was achieved by the robots accumulating on one side of the rod and pushing against it. For longitudinal transport, the robots have very little area against which forces can be applied to the rod because of its slender shape. However, the robots learned to slide along the surface of the rod and apply forces in the direction of the long axis of the rod via friction. Although inefficient, this sub-strategy enabled the microrobot swarm to control the third motion mode of the rod and features an impressive example of how the MARL algorithm can exploit environmental details to achieve its task.

Fig. 5D shows experimental snapshots of the robots, the rod and the target area during one transport episode in which the trained model was evaluated. Fig. 5E shows the corresponding rod trajectory in terms of its angle $\alpha$ and distance $D$ to the target (black trajectory). Initially, the rod is located at one of the ends of the target and is oriented orthogonally to it. The robots started by transporting it transversally toward the center of the target without altering the rod orientation. Only when the rod reached a distance of approximately 25 μm from the center of the target did the robots start to rotate it to align it with the target. In different post-training episodes with similar initial rod placements, we find that the microrobots used the same sequence of sub-strategies. This was also confirmed by 200 episodes carried out with a trained model in the virtual environment. We rationalize this sequence by considering that transporting the rod transversally is a very effective way to move it over large distances since the robots can apply forces on a large area. After placing the rod close to the target center, the robots only have to rotate it in the second half of the episode to match the target orientation. If we imagine a second sequence in which the robots first align the rod with the target and then fully move it into the target area, this would involve transporting the rod longitudinally, which is much less efficient (see Fig. 5F). In a scenario where the rod was initially placed in one line with the target, the microrobots even reliably misaligned the rod from the target first, which allowed for more efficient transport (see Fig. S11). In summary, the microrobot swarm is capable of solving the complex task of transporting a rod to a specific target location and orientation. The solution consists of several sub-strategies, that the robots combine in an intricate way to avoid inefficient transport modes.

**Scalability and robustness**

The above examples demonstrate that different collective tasks can be achieved with the same microrobotic swarm by simply changing the global measure of performance. We used the collective rotation of the rod as a test problem to probe the robustness and scalability of this system. For both tests, we did not allow the system to adopt its strategy (i.e., no further training) for the altered condition. We chose this approach because, in the particular case of a microrobotic system deployed under different conditions, no adjustments to the strategy would be possible owing to the limited sensing and computing capabilities of the robots.

To test the robustness of the swarm against malfunctions in the steering or decision-making of the microrobots, we ran multiple rotation episodes in the experiment after training, during which we intentionally introduced randomized actions. In each time step, a fraction of the actions chosen by the robots were set to one of the four possible actions at random instead of executing the action chosen by the policy of the robot. We randomized the actions of a different set of robots in every time step (see Fig. 6A-B). Randomizing the actions of a fixed set of robots would make that set of robots diffuse away and separate the functioning from the malfunctioning robots, effectively changing only the size of the robot swarm. Fig. 6C shows the performance in the rotation task (given by the average angular velocity of the rod) against the fraction of malfunctions introduced to a robot swarm. For up to 20 % malfunctions, the performance effectively remained the same as that when no malfunctions were added. For a higher fraction of randomized actions, the performance decayed. However, even with half the actions randomized, the robot swarm managed to maintain 30 % of the initial performance. With a level of 70 % malfunctions or more, the robot swarm can no longer execute the task. The reason is that at this high level of randomized actions, the robots cannot navigate, but the noise makes them diffuse away from the rod (see Supplementary Movie S6). The results of the robustness test demonstrate that our microrobot swarm can not only handle the highly stochastic environment at the micron scale but also remains functional in the presence of considerable robotic errors.

To assess the scalability of the microrobot swarm, we altered the number of robots in a trained group. Again, we used rod rotation performance as an indicator of the effects of different group sizes on function (Fig. 6D). We find that the performance is highest for robot numbers close to the size in which the swarm was trained (approximately 35). For a total of 20 robots, the performance remained unaltered. The performance decreased for smaller groups because of the limited torque that a low number of robots can apply to the rod. However, even with as few as 9 microrobots, the swarm retained approximately half of the maximal performance (see Supplementary Movie S6). When the number of robots was increased to more than twice the number in which the group was trained, the performance decreased as well. We can rationalize the reduced performance in larger groups considering that only a certain number of robots can access the rod and effectively help to rotate it. The decrease in performance is due to the interactions between the robots, which can substantially change their swimming behavior (see Fig. 3). When robots form a dense cluster around the whole rod, this effectively hinders the robots that are in contact in their efforts and blocks the rotation of the rod. In simulations where we trained microrobot swarms with different numbers of robots, the performance continued to increase as the number of robots increased (see Fig. S12). This shows that with further training and adaptation of the strategy, the swarm can be scaled up to even larger robot numbers.

In contrast to the rotation task, where the performance is given for each time step, the performance in the transport task is only given as an average success rate over multiple episodes. For this reason, we were not able to investigate scalability in the transport task to the same detail as for rod rotation. We did, however, run multiple transport episodes with different numbers of robots (25, 35 and 55) in the experiment and found that the average success rate increases with the number of robots (see Fig. S13). This trend is most likely because more robots are able to push the rod to the target faster and therefore are more likely to succeed within the limited time of one episode. Upon increasing the time available in one episode, the success rate of smaller microrobot swarms increased.

**Multi object manipulation**

As outlined in the introduction, distributed control and the independent motion of each microrobot enable swarms to show more rich and complex motion patterns than when controlled by a global field. To demonstrate the potential that this control scheme opens up, we let the microrobot swarm rotate two and three rods simultaneously in demonstration experiments after training (see Fig. 7 and Supplementary Movie S7). For these experiments, we divided the swarm into teams that are each responsible for the manipulation of one rod. We did not need to alter any part of the control framework and could choose the rotation direction of each rod independently. This demonstration shows the potential for independently moving microrobots to open up with a distributed control framework.

## Discussion

In this work, we demonstrate the experimental realization of a microrobotic swarm system of up to 200 individually controlled, self-propelled microrobots. We implement a state-of-the-art Multi-Agent Reinforcement Learning algorithm to achieve fully decentralized control. This enables individual microrobots to act on the basis of very limited information and navigate their highly stochastic environment. Using counterfactual rewards, we can promote the development of coordination strategies that allow the group to perform collective tasks beyond the capability of the single microrobots. Furthermore, we can adapt the MARL framework to train for different tasks without redesigning the reward function of individual robots but by simply redefining the global measure of performance. We demonstrated that the microrobot swarm system can learn to collectively rotate a large rod in end-to-end training in experiments. In addition, we showed that these microrobots can perform targeted transport of the same rod to an arbitrary position and orientation with an optimized control strategy. Selecting (and alternating between) tasks at execution time is a matter of a simple switch in the software and does not require any recalibration or modification of the experimental apparatus. In demonstration experiments, we let the microrobot swarm rotate two and three rods simultaneously in independent rotation directions, highlighting the benefits of individually moving microrobots with decentralized control.

Notably, the MARL-trained microrobot swarm can handle very high levels of noise, both at the individual level and as a collective ensemble. As shown in Fig. 3, the motion of individual microrobots is subject to thermal noise, and the outcome of each action is stochastic. Interactions with other robots and rods introduce even more chaotic motion. Despite the imperfect control over the motion of the robots, the trained policy allows the robots to navigate well enough for them to push against specific parts of the rod to achieve their task. At the system level, the MARL-trained control policy has shown robustness in terms of changing the number of agents and the fraction of malfunctioning units randomly assigned to the microrobotic swarm. Although part of the robustness comes from the nature of decentralized swarm robotic systems, this level of system robustness can ensure successful implementation for many potentially useful application scenarios where high levels of noise and uncertainty are common in complicated environments.

Another aspect we would like to highlight is the collective performance despite the limited information available to each microrobot. It may be obvious to guess the optimal control strategy with a bird's eye view and global information about all the microrobots. From the point of view of one microrobot, however, even the simple task of rotating a rod is much more complex. With locally perceived information and only 10-pixel visual state inputs, how to drive these micro "bumper cars" to optimally push a large rod is not obvious. In this respect, the tasks we have demonstrated with MARL are much more complicated than tasks previously solved by microrobot systems (*43, 44*). We also want to emphasize the flexibility of RL algorithms with respect to their state inputs. The input to the policy, which is still a human-readable "image" (with 10 pixels) here, could be replaced by any type of sensor reading that carries comparable information about the global state but could be much more abstract. In this case, a MARL algorithm with an appropriate reward scheme can achieve what no manually designed control algorithm or human operator can achieve.

Furthermore, we find that the learned policy in the case of targeted transport naturally divides the task into smaller, more tractable stages. During each stage, the robots moved the rod in one of its motion modes, and the chosen sequence was optimized for efficiency. The automatic subdivision of tasks is a key feature for solving complex problems involving multiple time and length scales. We believe that our microrobotic swarm system experimentally demonstrates the capability of MARL using CR in this context, which paves the way for achieving collective intelligence in future microscopic swarms.

Realistic environments are expected to be more complex than the uniform environment considered in our experiments. To demonstrate the adaptability of our framework to more realistic environments, we performed demonstration simulations in which we train the microrobot swarm to rotate and transport a rod in the presence of impenetrable fixed obstacles (see Fig. S14 and Supplementary Movie S8). The sensing of the robots was achieved similarly to the sensing of other robots and the target by adding additional state inputs for the perception of obstacles. We find that the microrobots learn to navigate around obstacles to reach the rod and even move the rod around the obstacles. Interestingly, the microrobots even learn to exploit the structured environment and use an obstacle as a hinge to transport the rod more efficiently.

The experimental realization of our independently controlled microrobotic swarm system relies on lasers and optics in a modern laboratory setting. Owing to the shallow penetration depth of lasers and the propulsion mechanism, the proposed microrobotic systems may not be suitable for many medical applications, such as targeted drug delivery inside the human body (*45*). Nonetheless, this kind of setup can complete microscopic manipulation or assembly tasks in on-chip settings. However, our work mainly aims to provide a control framework for robotic systems that are both highly resource-constrained and face noisy environments. Tiny robots are typically constrained in their computational and sensorial resources owing to limitations in hardware required for high-resolution sensors or communication with other robots or a central controlling unit. Moreover, miniaturized actuators cannot achieve the same precision that larger robots can and the interactions with other robots and the environment is more chaotic at small length scales. At the microscale, there are, to our knowledge, no fully autonomous robots yet, but we envision control frameworks such as ours to be used for microscopic robots that have very limited sensory information paired with very small computational capabilities. A simple neural network such as ours could be trained in, e.g., a virtual environment and then hardwired on a microscopic chip to be implemented on autonomous microscopic robots. The field of microrobots is rapidly evolving, and microfabrication and integration of microactuators, microsensors, and microcontrollers with existing CMOS circuits are becoming possible (*22*). The small size of the artificial neural network (3 hidden layers with 32, 16 and 16 nodes in this work) and the amount of power required to run such a policy is typically within reach for onboard implementation (*46*). We envision that a microrobot swarm with an end-to-end MARL control policy holds potential for many advanced lab-on-a-chip applications. For these potential applications, the ability to manipulate multiple objects simultaneously is especially valuable. Such systems could be used as microrobotic assembly lines that can collectively transport valuable biological samples for single-cell analysis (*47*), virus detection (*48*), and chemical and biological analysis platforms (*49*). Other potential applications are custom microrobot fabrication (*29*), the assembly of personalized drug-releasing capsules (*50*), and tissue engineering (*51*). Using holographic methods (*52, 53*), this type of microrobot swarm could even be adapted to three dimensions.

## Materials and Methods

### Laser-activated microswimmers

We fabricated light-responsive active particles from commercially available 6 μm silica spheres by depositing an 80 nm carbon layer on one hemisphere. These particles are then placed in a 1 cm × 3 cm × 200 μm sample cell containing a critical binary mixture of 26.8 wt% lutidine in water and kept at a temperature of 28.2 °C, close to its lower demixing point at 34 °C. When illuminated with a focused laser beam, the carbon-coated hemisphere heats up above the critical point. This causes localized heating, and the fluid around the particle separates, leading to self-propulsion of the particle (*34*). To achieve precise control of individual particle motion, we implemented a feedback loop. Images of the sample were acquired at a rate of 3–4 Hz (depending on the number of particles), and live image analysis and particle tracking were performed. We then used a 532 nm laser together with an acousto-optical deflector to selectively illuminate individual particles. By scanning over the particles at a rate of 100 kHz, we achieve quasi-continuous illumination of up to 200 robots simultaneously since relaxation of the demixing bubbles occurs on a time scale of approximately 100 ms (*36*). The setup is described in detail in the Optical setup section in the Supplementary Materials and Fig. S2. Due to the low intensities, the laser does not exert any optical forces. For active steering, we shift the laser position with respect to the particle position and orientation. Shifting it to the capped side of the particles with an average power of 2.7 µW ensures stable forward motion. For stationary actions, we use the same position but only 0.3 µW, which reduces the rotational diffusion of the robots. During rotation actions, two laser spots of different intensities (1.1 µW and 1.7 µW) were applied to the sides of the carbon cap. This creates a heat gradient across the carbon cap, causing anisotropic demixing and resulting in an active torque (*34, 54*). As a result, each swimmer can execute the four different actions forward, stay stationary, right turn and left turn. Each action takes 10 seconds, and the corresponding laser pattern is applied to each robot until it has traveled 6 µm during the forward actions or has rotated $\pi/5$ during a rotation action. After that, the laser pattern is set to stationary for the remaining time of the current action. During rotation actions, robots not only rotate but also translate around half their diameter (3 µm) on average. In addition to the driving forces applied via laser-induced demixing, the position and orientation of all the robots are subject to thermal noise. The robots have a diffusion coefficient of approximately 0.03 µm/s. The velocity of their Brownian motion is consequently on the same order of magnitude as the velocity achieved by laser control. Considering the robot diameter as the important length scale, the Péclet number of the system is approximately 100; considering the rod length as the important length scale, it is 2000. However, the Péclet number does not adequately represent the dynamics of our microrobots since the interaction between the robots and the rod introduces the largest part of the noise and is not considered in the diffusion coefficient of velocity. See Fig. S3 and Supplementary Movie S1 for more information about the motion of the robots. For training and evaluating our MARL algorithm, we use groups of 25--40 microrobots (with the exception of the scalability experiments). To ensure that no robots are lost and that no new particles diffuse into the field of view, the particles entering the feedback loop are propelled outward, and the robots inside the field of view that reach the boundary are rotated inward and propelled inward. Note that these boundary actions override the RL policy, so the trajectories of the particles reoriented due to the boundary conditions are not included in the training of the policy.

**Fabrication of ellipsoid rods**

The ellipsoid rod particles that the microrobots moved in the different tasks were fabricated via two-photon 3D printing using a NanoScribe (*37*).

**State input definition**

We implement the state inputs $o_i(s)$ as a crude approximation of vision for each robot. In the rotation task, the vision field is limited to 180° to the front of each robot. This vision field is divided into 5 cones of equal angles (36°). For targeted transport, we wanted the robots to be able to always detect both the rod and the target and therefore expanded the vision field to 360° by adding five more cones to the back of each robot, which yields 10 cones in total. For each cone, a scalar $o_i^l(s)$ is computed for each perceived species (other robots, the rod, and the target in the case of transport). In this calculation, the other robots $j$ each contribute with $\frac{\sigma}{|r_{ij}|}$, where $\sigma$ represents the robot diameter and where $|r_{ij}|$ is the distance between $i$ and $j$. The rod and target are both treated as lines of 60 particles of different species. This approach of inverse distance weighting of neighbors is a well-established way to replicate realistic vision (*55*) and approximates the coverage of the perceiving robot's horizon by the perceived species. To ensure stable results for short distances, the contribution of each perceived robot $j$ is bounded to 1. The expression for the state input of one cone $l$ for one robot $i$ is therefore

$$o_i^l(s) = min\left(\sum_{j \neq i}^{M_l} \frac{\sigma}{|r_{ij}|}, 1\right), \qquad (2)$$

where $M_l$ is the number of other robots (or rod or target particles) in cone $l$. For each robot, the contributions of all two (or three) perceived species are simply concatenated to one vector that is fed into the actor network. For more complex environments, the state inputs of the robots can be extended to any type of environmental factor, similar to the perception of the rod and the target. It is not necessary to change the MARL algorithm in any way since the additional information is processed by the ANNs directly. Fig. S14 shows the results of training for rotation and transport in the virtual environment in the presence of obstacles that the robots and the rod cannot penetrate.

**Reward definition**

To define the individual reward functions, we first define a global performance function $p$ reflecting the task the robot swarm is supposed to solve. With $P_t$, we denote the performance value obtained at time $t$. For the rotation of the ellipsoid, we simply choose the ellipsoid's angular velocity $p = \omega$ as the performance function. Since we reward the agents at discrete timesteps, we measure the instantaneous angular velocity $\omega_t$ at time $t$ as the absolute value of the difference in the ellipsoid's orientation $\theta$ before and after the last timestep and therefore

$$P_t = \theta_t - \theta_{t-1} \text{ V}. \qquad (3)$$

Note that we do not introduce a preferred direction of rotation in this way. For targeted transport, we define the global performance on the basis of a "potential" function $v$ that is given by the mean distance between 60 virtual segments $k$ of the ellipsoid and the target

$$v = \frac{1}{60} \sum_{k=1}^{60} d_k \qquad (4)$$

(for an illustration, see Fig. 5B), since only potential-based rewards do not distort tasks (*56*). Because the orientation of the ellipsoid is defined only between 0° and 180°, we always consider both directions of numbering the ellipsoid segments and choose the direction that yields the lower value of $v$. The swarm's performance $P_t$ at each timestep is then defined as the negative change in the potential $v$ during the last time step:

$$P_t = V_{t-1} - V_t. \qquad (5)$$

As mentioned in the introduction, rewarding single agents in a fully cooperative MARL setup (in a Dec-POMDP) is not trivial. Just giving each agent a team reward that reflects the global performance of the group leads to high variance in the reward signal of a single agent. We use an implementation of counterfactual rewards to estimate the contribution of each agent to the performance of the swarm. For rewarding agent $i$ at time $t$, we numerically estimate the group performance without $i$ ($P^v_{t \setminus i}$) during the last time step by performing a simulation starting from the previous state $s_{t-1}$ from which we exclude $i$ (we refer to the Numerical Simulations section for more details). For this re-simulation step, we do not add noise to the virtual environment. Adding noise here would increase the variance in the difference in the performance in the experiment, which is inherently noisy, and the re-simulations, which would in turn increase the variance in the reward signal and therefore result in worse training. We attempt to compensate for systematic deviations between the experimental and numerically estimated performances by rescaling all numerical performances $P^v_{t \setminus i}$ by the ratio between the real experimental performance $P_t$ and the numerical estimate $P^v_t$ containing all the agents:

$$\underline{P^v_{t \setminus i}} = P^v_{t \setminus i} \frac{P_t}{P^v_t}. \qquad (6)$$

The reward of agent $i$, $r_{t,i}$, is then given by the difference between the global performance in the experiment and the rescaled numerical performance without the robot $i$

$$r_{t,i} = \beta \left( P_t - \underline{P^v_{t \setminus i}} \right) \qquad (7)$$

where $\beta$ is a constant factor chosen such that the absolute values of the rewards mostly range from 0--1. The task of transporting the ellipsoid to the target is a truly episodic task; therefore, we gave all the agents a final reward of 500 if the robot swarm accomplished the task before the end of the episode. We set the success condition as the mean ellipsoid and target segment distance $v$ to be less than 8 μm.

## Supplementary Materials

- Supplementary Materials and Methods
- Supplementary Figures S1–S15
- Supplementary Table 1

# References and Notes

1. K. J. Benoit-Bird, W. W. Au, Cooperative prey herding by the pelagic dolphin, Stenella longirostris. *The Journal of the Acoustical Society of America* **125**, 125-137 (2009).
2. S. A. Ocko, A. Heyde, L. Mahadevan, Morphogenesis of termite mounds. *Proceedings of the National Academy of Sciences* **116**, 3379-3384 (2019).
3. E. O. Wilson, The sociogenesis of insect colonies. *Science* **228**, 1489-1495 (1985).
4. G. Vakanas, B. Krafft, Regulation of the number of spiders participating in collective prey transport in the social spider Anelosimus eximius (Araneae, Theridiidae). *Comptes rendus biologies* **327**, 763-772 (2004).
5. E. Osaba, J. Del Ser, A. Iglesias, X.-S. Yang, Soft Computing for Swarm Robotics: New Trends and Applications. *Journal of Computational Science* **39**, 101049 (2020).
6. M. M. Shahzad *et al.*, A Review of Swarm Robotics in a NutShell. *Drones* **7**, 269 (2023).
7. M. Schranz, M. Umlauft, M. Sende, W. Elmenreich, Swarm robotic behaviors and current applications. *Frontiers in Robotics and AI* **7**, 36 (2020).
8. S.-J. Chung, A. A. Paranjape, P. Dames, S. Shen, V. Kumar, A survey on aerial swarm robotics. *IEEE Transactions on Robotics* **34**, 837-855 (2018).
9. K. H. Petersen, N. Napp, R. Stuart-Smith, D. Rus, M. Kovac, A review of collective robotic construction. *Science Robotics* **4**, eaau8479 (2019).
10. E. Soria, F. Schiano, D. Floreano, Predictive control of aerial swarms in cluttered environments. *Nature Machine Intelligence* **3**, 545-554 (2021).
11. J. W. Romanishin, K. Gilpin, S. Claici, D. Rus, 3D M-Blocks: Self-reconfiguring robots capable of locomotion via pivoting in three dimensions. *2015 IEEE International Conference on Robotics and Automation (ICRA)*, 1925-1932 (2015).
12. J. Werfel, K. Petersen, R. Nagpal, Designing collective behavior in a termite-inspired robot construction team. *Science* **343**, 754-758 (2014).
13. F. Berlinger, M. Gauci, R. Nagpal, Implicit coordination for 3D underwater collective behaviors in a fish-inspired robot swarm. *Science Robotics* **6**, eabd8668 (2021).
14. M. Kouzehgar, M. Meghjani, R. Bouffanais, in *Global Oceans 2020: Singapore–US Gulf Coast*. (IEEE, 2020), pp. 1-8.
15. T.-Y. Huang, H. Gu, B. J. Nelson, Increasingly intelligent micromachines. *Annual Review of Control, Robotics, and Autonomous Systems* **5**, 279-310 (2022).
16. L. Yang *et al.*, A survey on swarm microrobotics. *IEEE Transactions on Robotics* **38**, 1531-1551 (2021).
17. J. Yan *et al.*, Reconfiguring active particles by electrostatic imbalance. *Nature materials* **15**, 1095-1099 (2016).
18. A. Bricard, J.-B. Caussin, N. Desreumaux, O. Dauchot, D. Bartolo, Emergence of macroscopic directed motion in populations of motile colloids. *Nature* **503**, 95-98 (2013).
19. H. Xie *et al.*, Reconfigurable magnetic microrobot swarm: Multimode transformation, locomotion, and manipulation. *Science robotics* **4**, eaav8006 (2019).
20. J. Yu, B. Wang, X. Du, Q. Wang, L. Zhang, Ultra-extensible ribbon-like magnetic microswarm. *Nature communications* **9**, 3260 (2018).
21. S. Martel, M. Mohammadi, in *2010 IEEE international conference on robotics and automation*. (IEEE, 2010), pp. 500-505.
22. M. Z. Miskin *et al.*, Electronically integrated, mass-manufactured, microscopic robots. *Nature* **584**, 557-561 (2020).
23. X. Dong *et al.*, Toward a living soft microrobot through optogenetic locomotion control of Caenorhabditis elegans. *Science Robotics* **6**, eabe3950 (2021).
24. M. Han *et al.*, Submillimeter-scale multimaterial terrestrial robots. *Science Robotics* **7**, eabn0602 (2022).
25. M. Kim, A. Yu, D. Kim, B. J. Nelson, S. H. Ahn, Multi-Agent Control of Laser-Guided Shape-Memory Alloy Microrobots. *Advanced Functional Materials*, 2304937 (2023).

26. F. N. Piñan Basualdo, A. Bolopion, M. Gauthier, P. Lambert, A microrobotic platform actuated by thermocapillary flows for manipulation at the air-water interface. *Science robotics* **6**, eabd3557 (2021).
27. F. A. Lavergne, H. Wendehenne, T. Bäuerle, C. Bechinger, Group formation and cohesion of active particles with visual perception-dependent motility. *Science* **364**, 70-74 (2019).
28. C. Bechinger *et al.*, Active particles in complex and crowded environments. *Reviews of Modern Physics* **88**, 045006 (2016).
29. Y. Alapan, B. Yigit, O. Beker, A. F. Demirörs, M. Sitti, Shape-encoded dynamic assembly of mobile micromachines. *Nature materials* **18**, 1244-1251 (2019).
30. S. Tasoglu, E. Diller, S. Guven, M. Sitti, U. Demirci, Untethered micro-robotic coding of three-dimensional material composition. *Nature communications* **5**, 3124 (2014).
31. J. Li, L. Li, S. Zhao, Predator–prey survival pressure is sufficient to evolve swarming behaviors. *New Journal of Physics* **25**, 092001 (2023).
32. M.-A. Blais, M. A. Akhloufi, Reinforcement learning for swarm robotics: An overview of applications, algorithms and simulators. *Cognitive Robotics*, (2023).
33. Y. Wang, M. Damani, P. Wang, Y. Cao, G. Sartoretti, Distributed reinforcement learning for robot teams: A review. *Current Robotics Reports* **3**, 239-257 (2022).
34. J. R. Gomez-Solano *et al.*, Tuning the motility and directionality of self-propelled colloids. *Sci Rep* **7**, 14891 (2017).
35. T. Bäuerle, A. Fischer, T. Speck, C. Bechinger, Self-organization of active particles by quorum sensing rules. *Nat Commun* **9**, 3232 (2018).
36. J. R. Gomez-Solano, S. Roy, T. Araki, S. Dietrich, A. Maciołek, Transient coarsening and the motility of optically heated Janus colloids in a binary liquid mixture. *Soft Matter* **16**, 8359-8371 (2020).
37. Y. Liu *et al.*, Structural color three-dimensional printing by shrinking photonic crystals. *Nature communications* **10**, 4340 (2019).
38. P. Sunehag *et al.*, Value-decomposition networks for cooperative multi-agent learning. *arXiv preprint arXiv:1706.05296*, (2017).
39. D. H. Wolpert, K. Tumer, Optimal payoff functions for members of collectives. *Advances in Complex Systems* **4**, 265-279 (2001).
40. M. N. Popescu, W. E. Uspal, A. Dominguez, S. Dietrich, Effective interactions between chemically active colloids and interfaces. *Accounts of chemical research* **51**, 2991-2997 (2018).
41. B. Liebchen, A. K. Mukhopadyay, Interactions in Active Colloids. *Journal of Physics: Condensed Matter*, (2021).
42. P. Stengele, A. Lüders, P. Nielaba, Capture and transport of rod-shaped cargo via programmable active particles. *Scientific Reports* **13**, 15071 (2023).
43. S. Muiños-Landin, A. Fischer, V. Holubec, F. Cichos, Reinforcement learning with artificial microswimmers. *Science Robotics* **6**, eabd9285 (2021).
44. L. Yang *et al.*, Autonomous environment-adaptive microrobot swarm navigation enabled by deep learning-based real-time distribution planning. *Nature Machine Intelligence* **4**, 480-493 (2022).
45. B. J. Nelson, S. Pané, Delivering drugs with microrobots. *Science* **382**, 1120-1122 (2023).
46. M. Verhelst, B. Moons, Embedded deep neural network processing: Algorithmic and processor techniques bring deep learning to iot and edge devices. *IEEE Solid-State Circuits Magazine* **9**, 55-65 (2017).
47. E. Shojaei-Baghini, Y. Zheng, Y. Sun, Automated micropipette aspiration of single cells. *Annals of biomedical engineering* **41**, 1208-1216 (2013).
48. H. Lin *et al.*, Ferrobotic swarms enable accessible and adaptable automated viral testing. *Nature* **611**, 570-577 (2022).
49. Y. Zhu, Y.-X. Zhang, L.-F. Cai, Q. Fang, Sequential operation droplet array: an automated microfluidic platform for picoliter-scale liquid handling, analysis, and screening. *Analytical chemistry* **85**, 6723-6731 (2013).

50. S. Y. Chin *et al.*, Additive manufacturing of hydrogel-based materials for next-generation implantable medical devices. *Science robotics* **2**, eaah6451 (2017).
51. R. Xie *et al.*, Magnetically driven formation of 3D freestanding soft bioscaffolds. *Science Advances* **10**, eadl1549 (2024).
52. X. Luo *et al.*, Real-Time 3D Tracking of Multi-Particle in the Wide-Field Illumination Based on Deep Learning. *Sensors* **24**, 2583 (2024).
53. J. E. Curtis, B. A. Koss, D. G. Grier, Dynamic holographic optical tweezers. *Optics communications* **207**, 169-175 (2002).
54. T. Bäuerle, R. C. Löffler, C. Bechinger, Formation of stable and responsive collective states in suspensions of active colloids. *Nat Commun* **11**, 2547 (2020).
55. S. V. Viscido, M. Miller, D. S. Wethey, The dilemma of the selfish herd: the search for a realistic movement rule. *Journal of theoretical biology* **217**, 183-194 (2002).
56. A. Y. Ng, D. Harada, S. Russell, Policy invariance under reward transformations: Theory and application to reward shaping. *Icml* **99**, 278-287 (1999).
57. M. Abadi *et al.*, *TensorFlow: Large-Scale Machine Learning on Heterogeneous Systems*.  (2015).
58. H. Yuen, J. Princen, J. Illingworth, J. Kittler, Comparative study of Hough transform methods for circle finding. *Image and vision computing* **8**, 71-77 (1990).
59. J. Schulman, S. Levine, P. Abbeel, M. Jordan, P. Moritz, in *International conference on machine learning*. (PMLR, 2015), pp. 1889-1897.
60. J. Schulman, F. Wolski, P. Dhariwal, A. Radford, O. Klimov, Proximal Policy Optimization Algorithms. 2017.
61. K. Zhang, Z. Yang, T. Başar, Multi-agent reinforcement learning: A selective overview of theories and algorithms. *Handbook of reinforcement learning and control*, 321-384 (2021).

## Acknowledgments

We thank Robert Löffler for very insightful discussions and Jakob Steindl for the microfabrication of ellipsoid rod particles.

## Funding

This project was funded by the DFG Centre of Excellence 2117, Germany "Centre for the Advances Study of Collective Behaviour", ID: 422037984. H.G. would like to acknowledge support from the SNSF Postdoc Mobility Grant (203203).

## Author contributions

V.L.H. conceptualization, performed experiments, wrote experiment code and simulation environment code, performed simulations, analyzed data, wrote the original draft preparation, and visualized the data. E.P. wrote the MARL code and simulation environment code, discussion of the results, writing – original draft preparation. H.G. discussion of the results, writing – original draft, visualization. C.B. conceptualization, discussion of the results, writing – original draft preparation.

## Competing Interests

The authors declare that there are no competing interests.

## Data and materials availability

All the data needed to evaluate the conclusions in the paper is presented in the paper or the Supplementary Materials. All used simulation and MARL code are publicly available at https://github.com/vheuthe/microbot_rl and can be cited at https://doi.org/10.5281/zenodo.13380516.

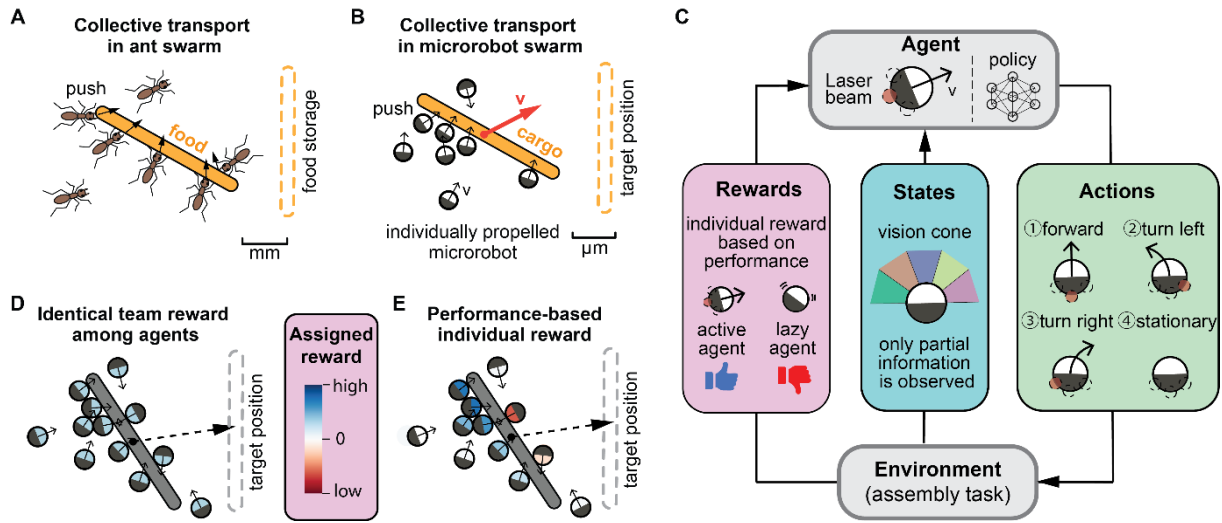

**Figure 1: Nature-inspired collective transport of large cargo in an individually controlled microrobot system.** (**A**) A group of ants collectively transports a large cargo object to a target location. (**B**) Microrobot swarms can perform a similar task. Each microrobot is individually controlled to swim forward and push the cargo toward the target. (**C**) Schematic diagram of the implementation of reinforcement learning (RL) in a laser-controlled microrobotic system. An agent, consisting of a microrobot and its policy, explores its environment by selecting actions on the basis of its own state inputs. After one time step, the environment gives back a reward, which rates how well the agent performs. The policy is updated to facilitate productive actions (yielding high rewards) and avoid futile actions (yielding low rewards). When using Multi-Agent Reinforcement Learning (MARL) to optimize the strategy of the robots, one can (**D**) give a joint reward to all the robots in the swarm or (**E**) reward each robot according to its contribution. Individually assigned rewards reduce the noise in the reward signal and facilitate training.

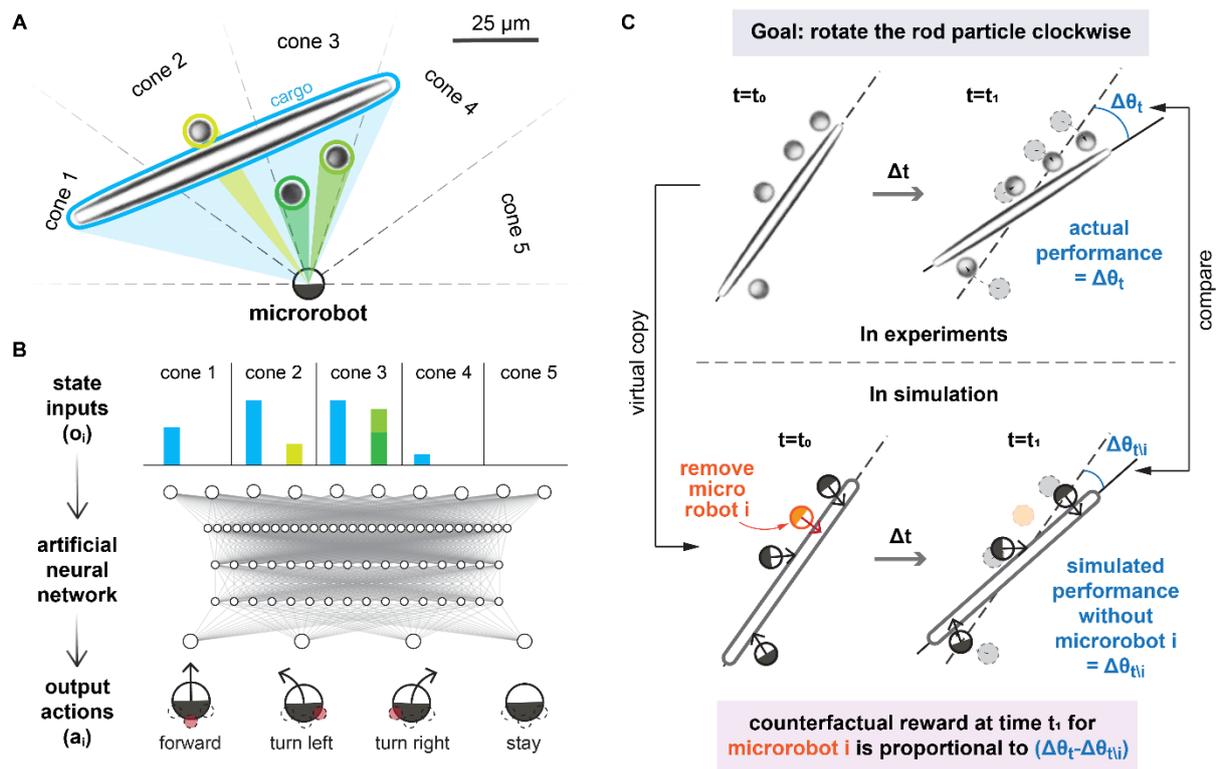

**Figure 2: Illustration of the system**. (**A**) Multiple 6 μm microrobots and the 100 μm rod with which they interact. The dashed lines indicate the detection cones of one robot. The robots and the rod contribute to the state input of the cone in which they are located, which is inversely weighted by their distance. (**B**) State inputs of the focal robot in (A) as a bar diagram used as the input to an artificial neural network (ANN) that decides on an action for this robot. The four possible actions are forward, turn left, turn right and stay stationary. We use a Multi-Agent Reinforcement Learning algorithm to optimize the control ANN so that the microrobot swarm can collectively complete a task. (**C**) For training, we use a counterfactual reward scheme. In the example of the task of rotating the rod, this kind of reward for one robot $i$ is calculated from the difference in the real angular velocity of the rod and the angular velocity of the rod in a virtual environment without the robot $i$ (for more details, refer to the Reward definition section in the Supplementary Materials). $\Delta\theta_t$ refers to the change in the angle of the rod, and the subscript $\setminus i$ means "without robot $i$".

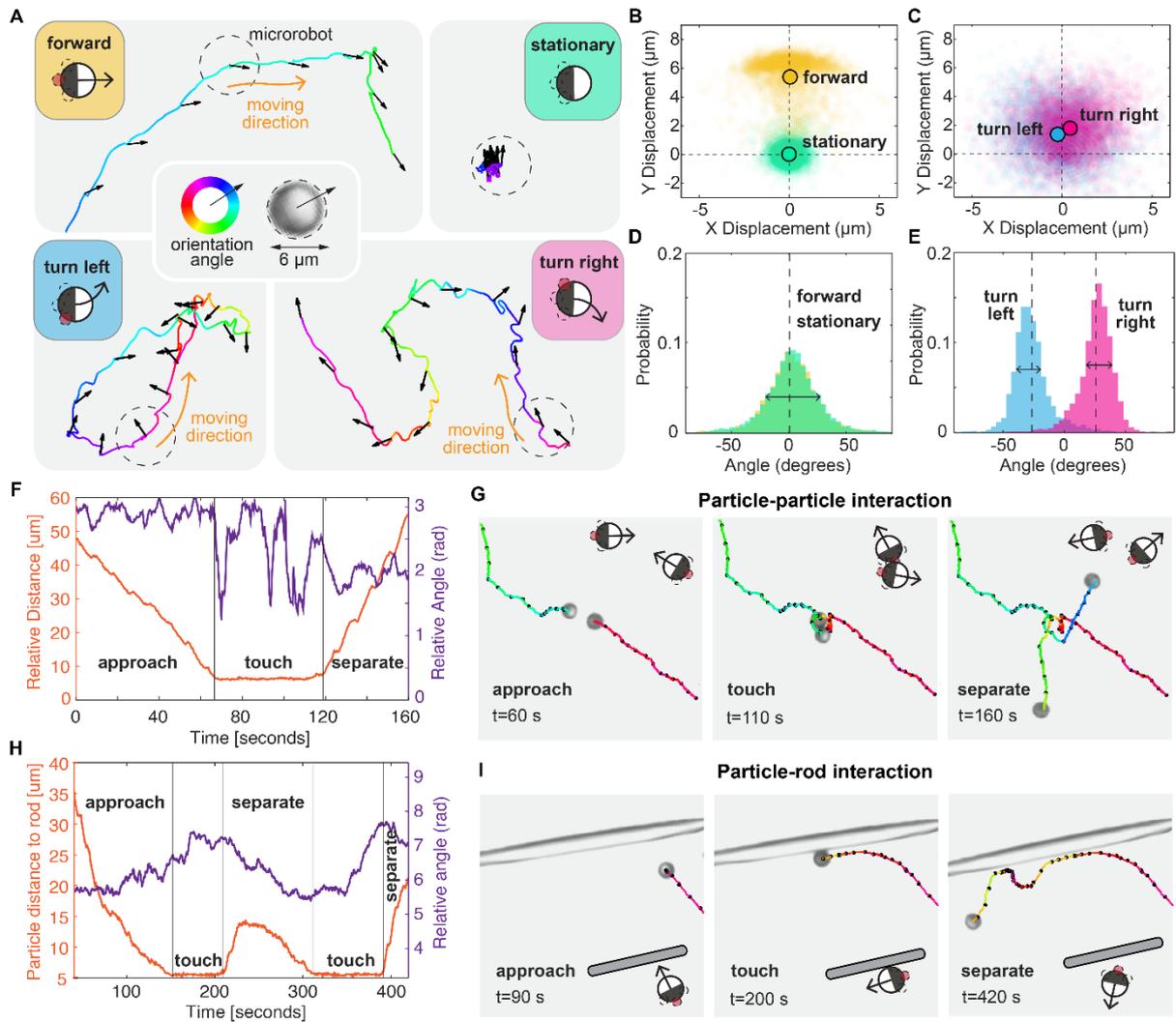

**Figure 3: Characterization of the individual robots.** (**A**) Trajectories of individual robots repeatedly performing one of the four possible actions forward, stay stationary, turn right or turn left. The robot orientation is shown as the color of the trajectories and indicated by arrows at the beginning of each action. (**B-E**) Statistical evaluation of the robots' motion during single actions. Upper row: x- and y-displacement during one action as scatter plots (B-C) and probability distributions (D-E). During a forward action, a robot moves about one diameter (6 µm) in the direction of its orientation. During rotation actions, a robot moves on average less than half a diameter (3 µm) forward. However, the direction of displacement is far more stochastic than during forward actions. Lower row: histograms of the change in the robot's orientation during one action. For the forward and stationary actions, the robots on average maintain their orientation, but the distribution is broad, with a FWHM of approximately 30° due to rotational diffusion. During one rotation action, the robots rotate by 36° (the angle of one detection cone) to the right and left, with an FWHM of approximately 17° (see Fig. S3 for more information). (**F**) Time evolution of the distance between two colliding robots in (**G**) (orange) and their relative angle. The microswimmers adhere for a short time, during which the fluctuations in orientation are largely enhanced, before the robots detach again. This type of complex interaction makes it difficult to model microrobotic systems, and needs to be considered by any controlling scheme. (**H-I**) A robot colliding with a rod shows similar behavior. This phenomenon complicates the control of robots that are pushing against the rod.

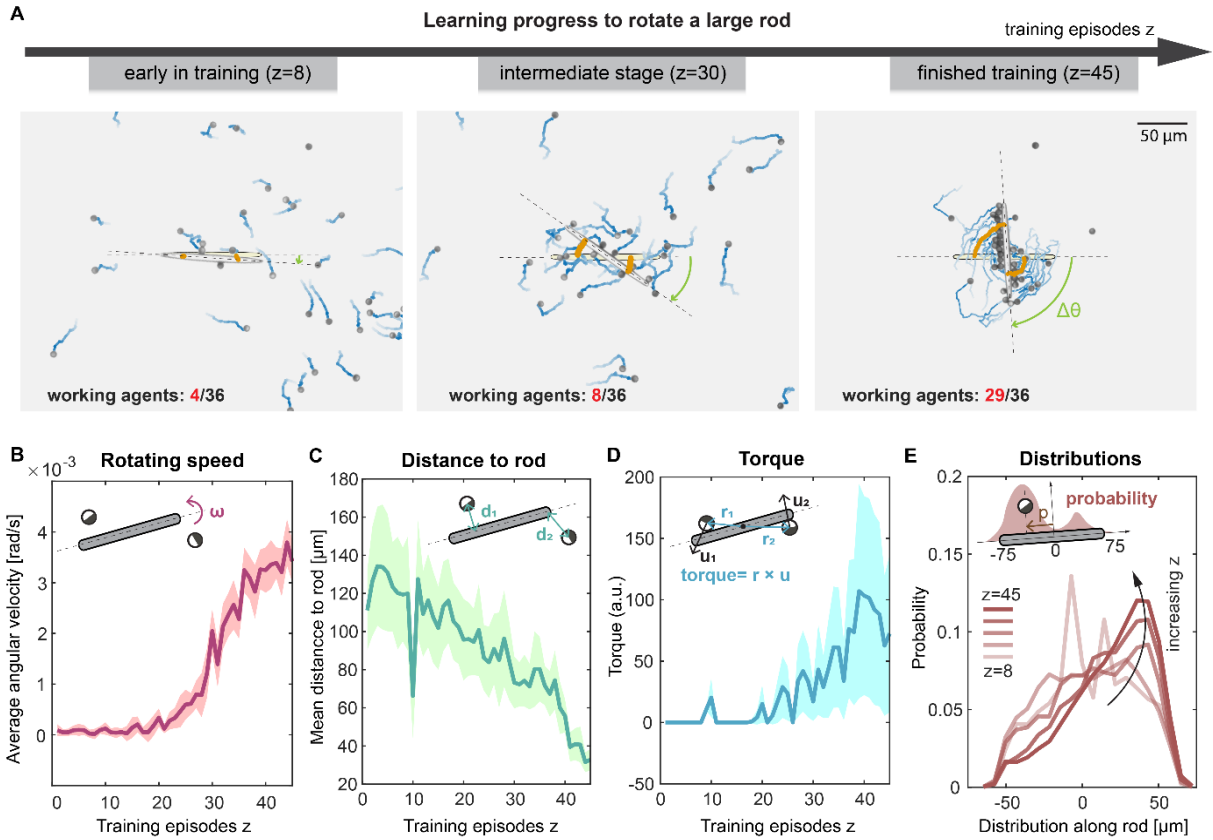

**Figure 4: Training the microrobot swarm to rotate a large rod particle.** (**A**) Snapshots of the robot's behavior during training, with the robots' trajectories in blue and the trajectories of two points on the rod in orange. (**B-E**) Evolution of the mean angular velocity of the rod (B), average distance of the robots to the rod (C), average torque applied to the rod (D) and the robot distribution along the rod (E) together with definitions of the evaluated quantities. The shaded areas represent the standard deviation. During the initial phase of the training, the robots mainly learn to navigate toward the rod (indicated by a decreasing average distance to the rod). After approximately 20 episodes, the robots learn how to apply a torque to the rod to rotate it. Their strategy involves the formation of two clusters at the rod ends that push against the rod from opposite directions (reflected in an asymmetric distribution of the robots along the rod).

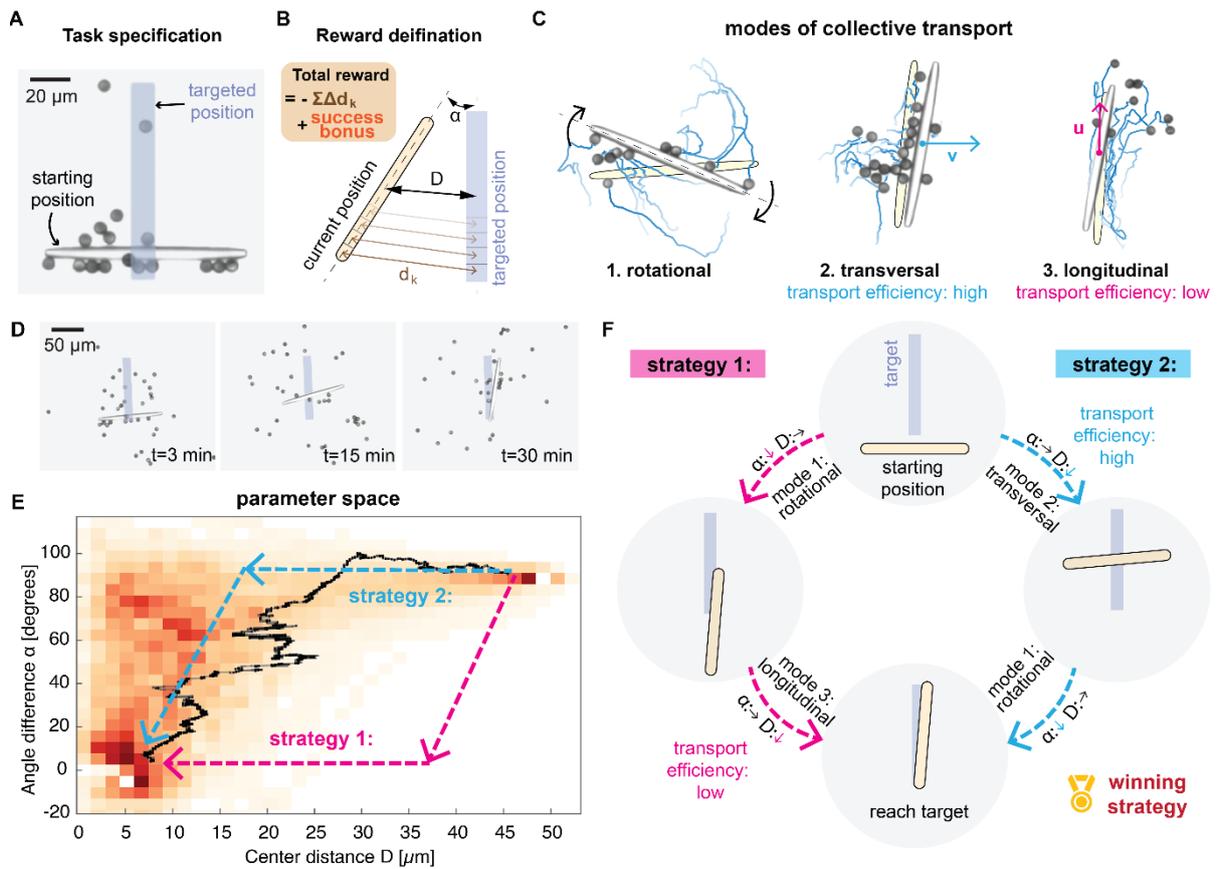

**Figure 5: Targeted transport task:** (**A**) The robots are tasked with placing the rod inside a predefined target region (marked in blue). (**B**) We use the negative change in the pairwise distance $d_k$ between 60 rod segments and 60 corresponding target segments $k$ as a measure of performance. On this basis, we compute counterfactual rewards for each robot and give all the robots a success bonus after the task was achieved. (**C**) The following strategies are used by the robot swarm to control the three modes of rod motion in the trained swarms: pushing against its ends as in the rotation task for rotating the rod, pushing against the side of the rod to move it transversally, and sliding along the rod to move it longitudinally. (**D**) Snapshots of the rod being transported toward the target after training together with a graph representing the trajectory of the rod relative to the target in terms of center distance and angle difference (**E**). The robots first move the rod to the center of the target and only then rotate it to align with the target area. Orange: probability density map of 200 simulated episodes starting in the same initial configuration with a trained model, showing that this behavior is a consistent strategy. (**F**) Comparison of the experimentally discovered strategy (blue) with the contrary strategy (red), in which the robot swarm first rotates the rod and then transports it. The experimentally found strategy involves transverse transport, whereas the contrary strategy involves the rather inefficient longitudinal transport mode.

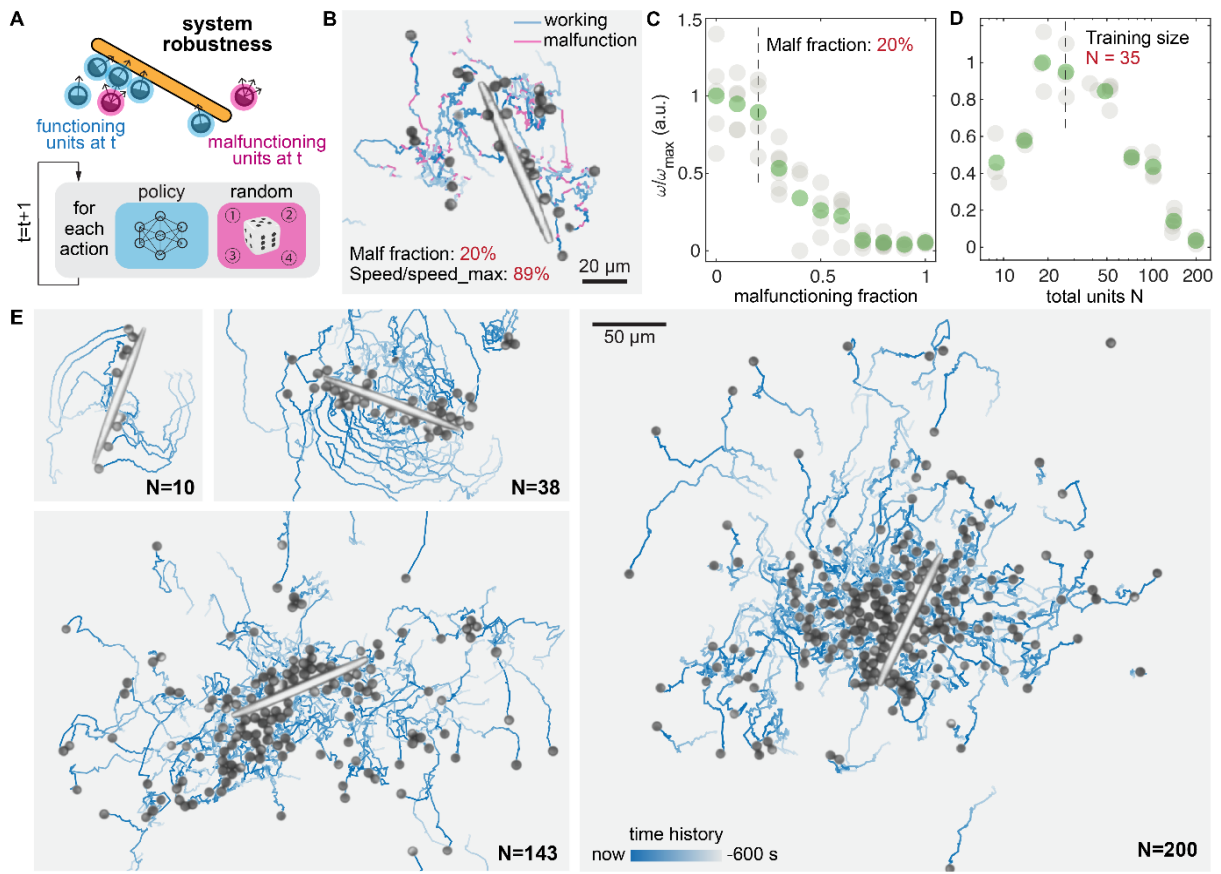

**Figure 6: Scalability and robustness in the rotation task:** (**A**) Schematic illustration of the introduction of malfunctions to a trained model during the rotation task: a part of the robot group is assigned a random action instead of the one its policy picked. The set of malfunctioning robots is re-chosen at every step to prevent the separation of malfunctioning robots from functioning robots. (**B**) The malfunctions introduce additional noise to the trajectories, marked in pink here. (**C**) Normalized rotation performance of a robot swarm with different numbers of introduced malfunctions without additional training (gray: individual experiments, green: mean value). For up to 20 % of randomized actions, the performance remains mostly unaltered. For very high fractions of malfunctions, the microrobots solely diffuse, and the performance decreases. (**D**) Normalized rotation performance $\omega/\omega_{max}$ of a robot swarm whose size is changed after training (gray: individual experiments, green: mean value). The performance is maximal around the swarm size, which is used for training. For very low and very high robot numbers, the performance is reduced, but for a range of one order of magnitude, half of the original performance is retained. (**E**) Experimental snapshots of a robot swarm whose size is changed after training, corresponding to the graph in (D).

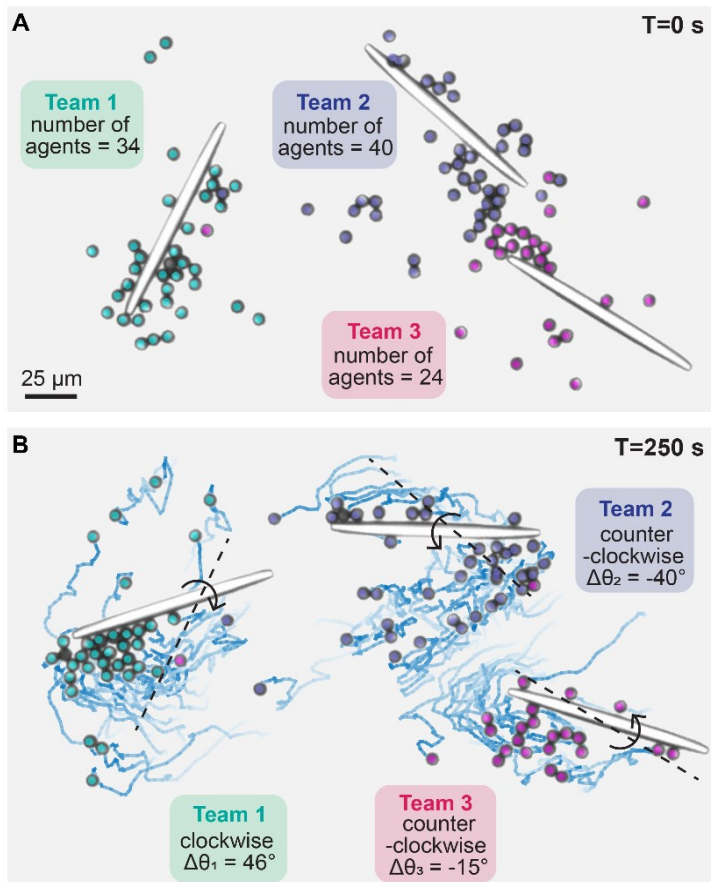

**Figure 7: Demonstration of multi-object manipulation with arbitrary rotating directions**: (**A**) Snapshot of three microrobot teams (colored) with three rods. (**B**) Snapshot of the three teams after 250 s with microrobot trajectories. The dashed lines indicate the initial orientation of the rods as in (A) and the arrows the rotation direction. The microrobot swarm can manipulate three rods simultaneously with independent rotation directions. For a video of multi-object manipulation see Supplementary Movie S7.

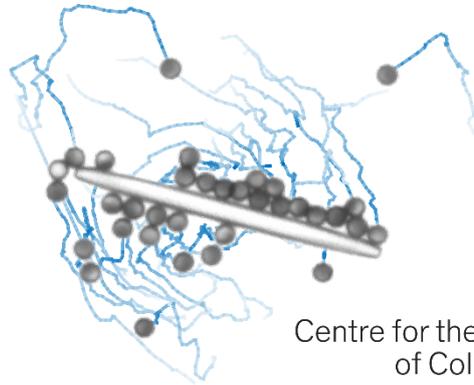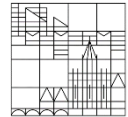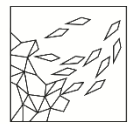

**Movie 1: Summary video of counterfactual rewards promote collective transport using individually controlled swarm microrobots**. This video summarizes the background, research motivation, main methods, and experimental results of the collective functions of microrobots powered by multi-agent reinforcement learning.